\documentclass[11pt,a4paper]{article}

\usepackage{times}
\usepackage{latexsym}

\usepackage[utf8]{inputenc} 
\usepackage[T1]{fontenc}    
\usepackage[english]{babel}
\usepackage{amssymb, amsmath}
\usepackage{url}            
\usepackage{relsize}        
\usepackage[skip=5pt]{caption}
\usepackage{booktabs}       
\usepackage{amsfonts}       
\usepackage{nicefrac}       
\usepackage{microtype}      
\usepackage[pdftex]{graphicx}
\usepackage{float}
\usepackage[numbers,sort,compress]{natbib}
\usepackage{hyperref}       
\usepackage[hyperref]{myacl}

\aclfinalcopy 

\restylefloat{figure}
\title{Discovering topics in text datasets by visualizing relevant words}

\author{Franziska Horn$^1$, Leila Arras$^2$, Gr{\'e}goire Montavon$^1$,\\ \textbf{Klaus-Robert M{\"u}ller$^{1,3}$, and Wojciech Samek$^2$}\\
  $^1$Machine Learning Group, Technische Universit\"at Berlin, Berlin, Germany \\
  $^2$Machine Learning Group, Fraunhofer Heinrich Hertz Institute, Berlin, Germany \\
  $^3$Department of Brain and Cognitive Engineering, Korea University, Seoul, Korea \\
  {\tt franziska.horn@campus.tu-berlin.de}}

\date{}

\begin{document}
\setlength{\abovedisplayskip}{4pt}
\setlength{\belowdisplayskip}{4pt}
\setlength{\intextsep}{9pt plus 3pt minus 2pt}
\setlength{\textfloatsep}{9pt plus 3pt minus 2pt}
\setlength{\floatsep}{5pt plus 3pt minus 2pt}
\setlength{\dblfloatsep}{7pt plus 3pt minus 2pt}
\setlength{\dbltextfloatsep}{12pt plus 3pt minus 2pt}

\maketitle

\begin{abstract}
When dealing with large collections of documents, it is imperative to quickly get an overview of the texts' contents. In this paper we show how this can be achieved by using a clustering algorithm to identify topics in the dataset and then selecting and visualizing relevant words, which distinguish a group of documents from the rest of the texts, to summarize the contents of the documents belonging to each topic. We demonstrate our approach by discovering trending topics in a collection of New York Times article snippets.
\end{abstract}

\section{Introduction}
Large, unstructured text datasets, e.g.~in the form of data dumps leaked to journalists, are becoming more and more frequent. To quickly get an overview of the contents of such datasets, tools for exploratory analysis are essential. 

We propose a method for extracting from a set of texts the relevant words that distinguish these documents from others in the dataset. By using the DBSCAN clustering algorithm \cite{ester1996dbscan}, the documents in a dataset can be grouped to reveal salient topics. We can then summarize the texts belonging to each topic by visualizing the extracted relevant words in word clouds, thereby enabling one to grasp the contents of the documents at a glance.

By identifying relevant words in clusters of recent New York Times article snippets, we demonstrate how our approach can reveal trending topics.

All tools discussed in this paper as well as code to replicate the experiments are available as an open source Python library.\footnote{\url{https://github.com/cod3licious/textcatvis}}

\subsection{Related work}
Identifying relevant words in text documents was traditionally limited to the area of feature selection, where different approaches were used to discard `irrelevant' features in an attempt to improve the classification performance by reducing noise as well as save computational resources \cite{forman2003extensive}. However, the primary objective here was not to identify words that best describe the documents belonging to certain clusters, but to identify features that are particularly uninformative in a classification task and can be disregarded. Other work was focused on selecting keywords for individual documents, e.g.~based on tf-idf variants \cite{lee2008news} or by using classifiers \cite{hulth2003improved,zhang2006keyword}. Yet, while these keywords might provide adequate summaries of single documents, they do not necessarily overlap with keywords found for other documents about this topic and therefore it is difficult to aggregate them to get an overview of the contents of a group of texts. Current tools available for creating word clouds as a means of summarizing a (collection of) document(s) mostly rely on term frequencies (while ignoring stopwords), possibly combined with part-of-speech tagging and named entity recognition to identify words of interest \cite{heimerl2014word,mcnaught2010using}. While an approach based on tf-idf features selects words occurring frequently in a group of documents, these words do not reliably distinguish the documents from texts belonging to other clusters \cite{horn2017exploring}. In more recent work, relevant features were selected using \emph{layerwise relevance propagation} (LRP) to trace a classifier's decision back to the samples' input features \cite{bach2015pixel,montavon2017methods}. This was successfully used to understand the classification decisions made by a convolutional neural network trained on a text categorization task and to subsequently determine relevant features for individual classes by aggregating the LRP scores computed on the test samples \cite{arras2016explaining,arras2016relevant}. While in classification settings LRP works great to identify relevant words describing different classes of documents, this method is not suited in our case as we are dealing with unlabeled data.

\section{Methods}
To get a quick overview of a text dataset, we want to identify and visualize the `relevant words' occurring in the collection of texts. We define relevant words as some characteristic features of the documents, which distinguish them from other documents. As the first step in this process, the texts therefore have to be preprocessed and transformed into feature vectors (Section~\ref{subsec:features}). While relevant words are supposed to occur often in the documents of interest, they should also distinguish them from other documents. When analyzing a whole dataset it is therefore most revealing to look at individual clusters and obtain the relevant words for each cluster, i.e.~find the features that distinguish one cluster (i.e.~topic) from another. To cluster the documents in a dataset, we use the DBSCAN algorithm (Section~\ref{subsec:clustering}). 

The relevant words for a cluster $c$ are identified by computing a relevancy score $r_c$ for every word $t_i$ (with $i=1...T$, where $T$ is the number of unique terms in the given vocabulary) and then the word clouds are created using the top ranking words. The easiest way to compute relevancy scores is to simply check for frequent features in a selection of documents. However, this does not necessarily produce features that additionally occur infrequently in other clusters. Therefore, we instead compute a score for each word indicating in how many documents of one cluster it occurs compared to other clusters (Section~\ref{subsec:relwords}).

\subsection{Preprocessing \& Feature extraction} \label{subsec:features}
All $N$ texts in a dataset are preprocessed by lowercasing and removing non-alphanumeric characters. Then each text is transformed into a \emph{bag-of-words} (BOW) feature vector $\mathbf{x}_k \in \mathbb{R}^T \;\forall k \in 1...N$ by first computing a normalized count, the \emph{term frequency} (tf), for each word in the text, and then weighting this by the word's \emph{inverse document frequency} (idf) to reduce the influence of very frequent but inexpressive words that occur in almost all documents (such as `and' and `the')~\cite{irbook, yang1997comparative}. The idf of a term $t_i$ is calculated as the logarithm of the total number of documents, $|N|$, divided by the number of documents which contain term $t_i$, i.e.
\begin{align*}
 \text{idf}\,(t_i) &= \log {|N|\over |\{k \in N\text{ : }t_i \in k\}|}.
 \end{align*}
The entry corresponding to the term $t_i$ in the tf-idf feature vector $\mathbf{x}_k$ of a document $k$ is then
\begin{align*}
\mathbf{x}_{ki} &= \text{tf}_k(t_i) \cdot \text{idf}\,(t_i).
\end{align*}
In addition to single terms, we are also considering meaningful combinations of two words (i.e.~bigrams) as features. However, to not inflate the feature space too much (since later, relevancy scores have to be computed for every feature), only distinctive bigrams are selected. This is achieved by computing a bigram score for every combination of two words occurring in the corpus similar as in \cite{mikolov2013distributed} and then selecting those with a score significantly higher than that of random word combinations. Further details can be found in the appendix of \cite{horn2017exploring}.

\subsection{Clustering} \label{subsec:clustering}
To identify relevant words summarizing the different topics in the dataset, the texts first have to be clustered. For this, we use \emph{density-based spatial clustering of applications with noise} (DBSCAN) \cite{ester1996dbscan}, a clustering algorithm that identifies clusters as areas of high density in the feature space, separated by areas of low density. This algorithm was chosen as it does not assume that the clusters have a certain shape (unlike e.g.~the k-means algorithm, which assumes spherical clusters) and it allows for noise in the dataset, i.e. does not enforce that all samples belong to a certain cluster.

DBSCAN is based on pairwise distances between samples and first identifies `core samples' in areas of high density and then iteratively expands a cluster by joining them with other samples, whose distance is below some user defined threshold. As the cosine similarity is a reliable measure of similarity for text documents, we compute the pairwise distances used in the DBSCAN algorithm by first reducing the documents' tf-idf feature vectors to 250 linear kernel PCA components to remove noise and create more overlap between the feature vectors~\cite{scholkopf1998nonlinear}, and then compute the cosine similarity between these vectors and subtract it from $1$ to transform it into a distance measure. As clustering is an unsupervised process, a value for the distance threshold has to be chosen such that the obtained clusters seem reasonable. In the experiments described below, we found that a minimum cosine similarity of $0.55$ to other samples in the cluster (i.e.~using a distance threshold of $0.45$) leads to texts about the same topic being grouped together.

We denote as $y_k$ the cluster that document $k$ was assigned to in the clustering procedure.

\subsection{Identifying relevant words} \label{subsec:relwords}
Relevant words for each cluster are identified by computing a relevancy score $r_c$ for every word $t_i$ and then selecting the highest scoring words.

\begin{figure}[!h]
  \centering
      \includegraphics[width=\columnwidth]{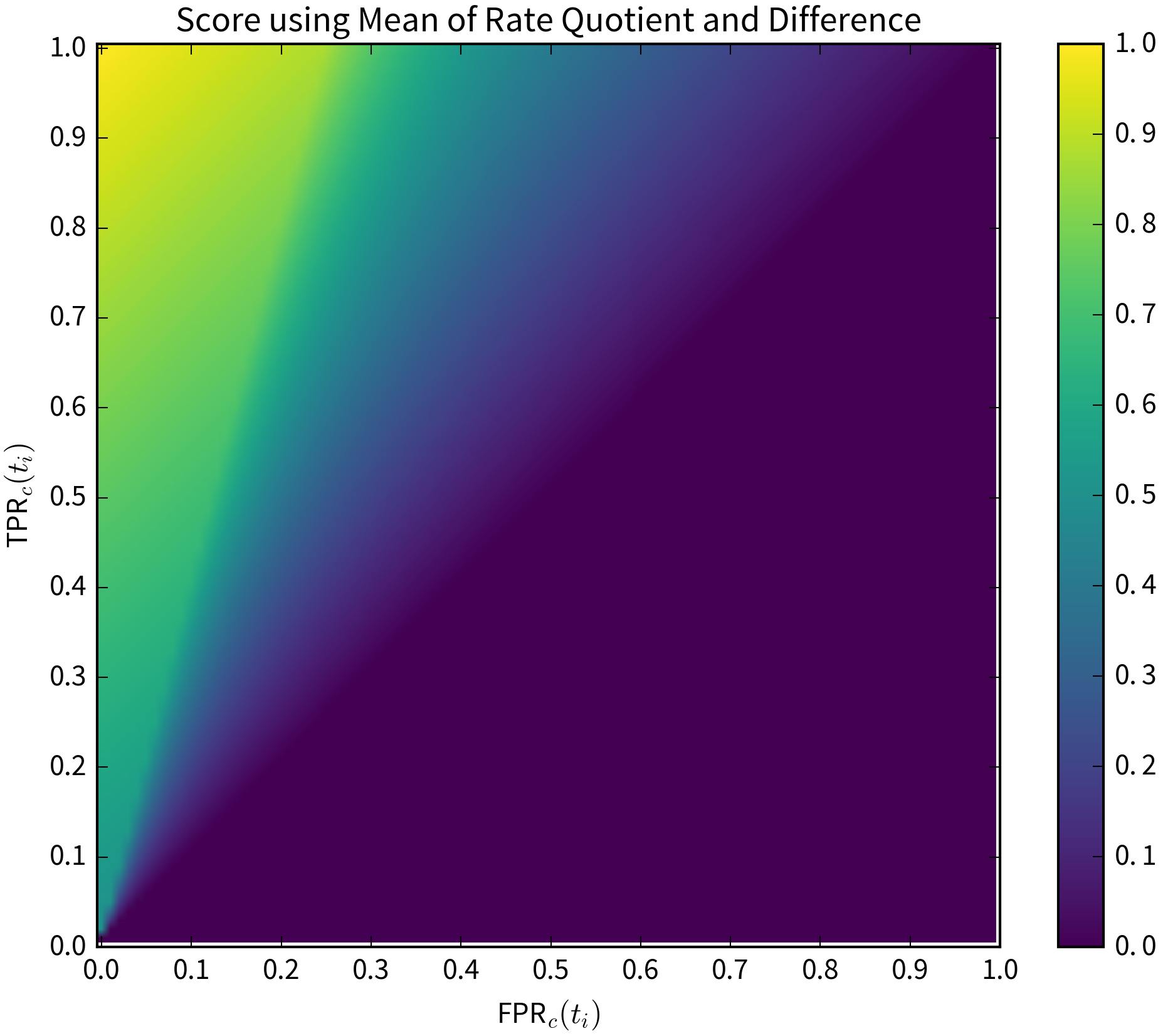}
  \caption{Relevancy score $r_c(t_i)$ depending on a word's TPR and FPR for a cluster.}
  \label{fig:distinctive_scores}
\end{figure}

We compute a score for each word depending on the number documents it occurs in from one cluster compared to the documents from other clusters. We call the fraction of documents from a target cluster $c$ that contain the word $t_i$ this word's true positive rate
\begin{align*}
\text{TPR}_c(t_i) = {|\{k\text{ : }y_k = c \wedge \text{tf}_{k}(t_i) > 0\}|\over |\{k\text{ : }y_k = c\}|}\,.
\end{align*}
Correspondingly, we can compute a word's false positive rate as the mean plus the standard deviation of the TPRs of this word for all other clusters:\footnote{We are not taking the maximum of the other clusters' TPRs for this word to avoid a large influence of a cluster with maybe only a few samples.}
\begin{align*}
\text{FPR}_c(t_i) =& \text{ mean}(\{\text{TPR}_l(t_i) : l \neq c\})\\ &+ \text{std}(\{\text{TPR}_l(t_i) : l \neq c\})\,.
\end{align*}
The objective is to find words that occur in many documents from the target cluster (i.e.~have a large $\text{TPR}_c(t_i)$), but only occur in few documents of other clusteres (i.e.~have a low $\text{FPR}_c(t_i)$). One way to identify such words would be to compute the difference between both rates, i.e.
\begin{align*}
r_c\_\text{diff}\,(t_i) = \max\{\text{TPR}_c(t_i) - \text{FPR}_c(t_i), 0\}\,,
\end{align*}
which is similar to traditional feature selection approaches \cite{forman2003extensive}.
However, while this score yields words that occur more often in the target cluster than in other clusters, it does not take into account the relative differences. For example, to be able to detect emerging topics in newspaper articles, we are not necessarily interested in words that occur often in today's articles and infrequently in yesterday's. Instead, we acknowledge that \emph{not most articles} published today will be written about some new event, only \emph{significantly more articles} compared to yesterday. Therefore, we propose instead a rate quotient, which gives a score of $1$ to every word that has a TPR about three times higher than its FPR:
\begin{align*}
r_c\_\text{quot}\,(t_i) &= {\min\{\max\{z_c(t_i),1\},4\}-1 \over 3},\\
 \text{with } z_c(t_i) &= {\text{TPR}_c(t_i) \over \max\{\text{FPR}_c(t_i), \epsilon\}}\,.
\end{align*}
While the rate quotient extracts relevant words that would otherwise go unnoticed, for a given FPR of $0.05$ it assigns the same score to words with a TPR of $0.3$ and a TPR of $1.0$. Therefore, to create a proper ranking amongst all relevant words, we take the mean of both scores to compute the final score,
\begin{align*}
r_c\,(t_i) & = 0.5\left(r_c\_\text{diff}\,(t_i) + r_c\_\text{quot}\,(t_i)\right),
\end{align*}
which results in the TPR-FPR relation shown in Fig.~\ref{fig:distinctive_scores}.

\section{Experiments \& Results}
To illustrate how the identified relevant words can help when exploring new datasets, we test the previously described methods on recent article snippets from the New York Times. The code to replicate the experiments is available online and includes functions to cluster documents, extract relevant words and visualize them in word clouds, as well as highlight relevant words in individual documents.\footnote{\url{https://github.com/cod3licious/textcatvis}}

To see if our approach can be used to discover trending topics, we are using newspaper article snippets from the week of President Trump's inauguration (Jan $16^{\text{th}}$-$22^{\text{nd}}$, 2017), as well as three weeks prior (including the last week of 2016), downloaded with the Archive API from New York Times.\footnote{\url{https://developer.nytimes.com/archive_api.json}}
\begin{figure}[!h]
  \centering 
      \includegraphics[width=\columnwidth]{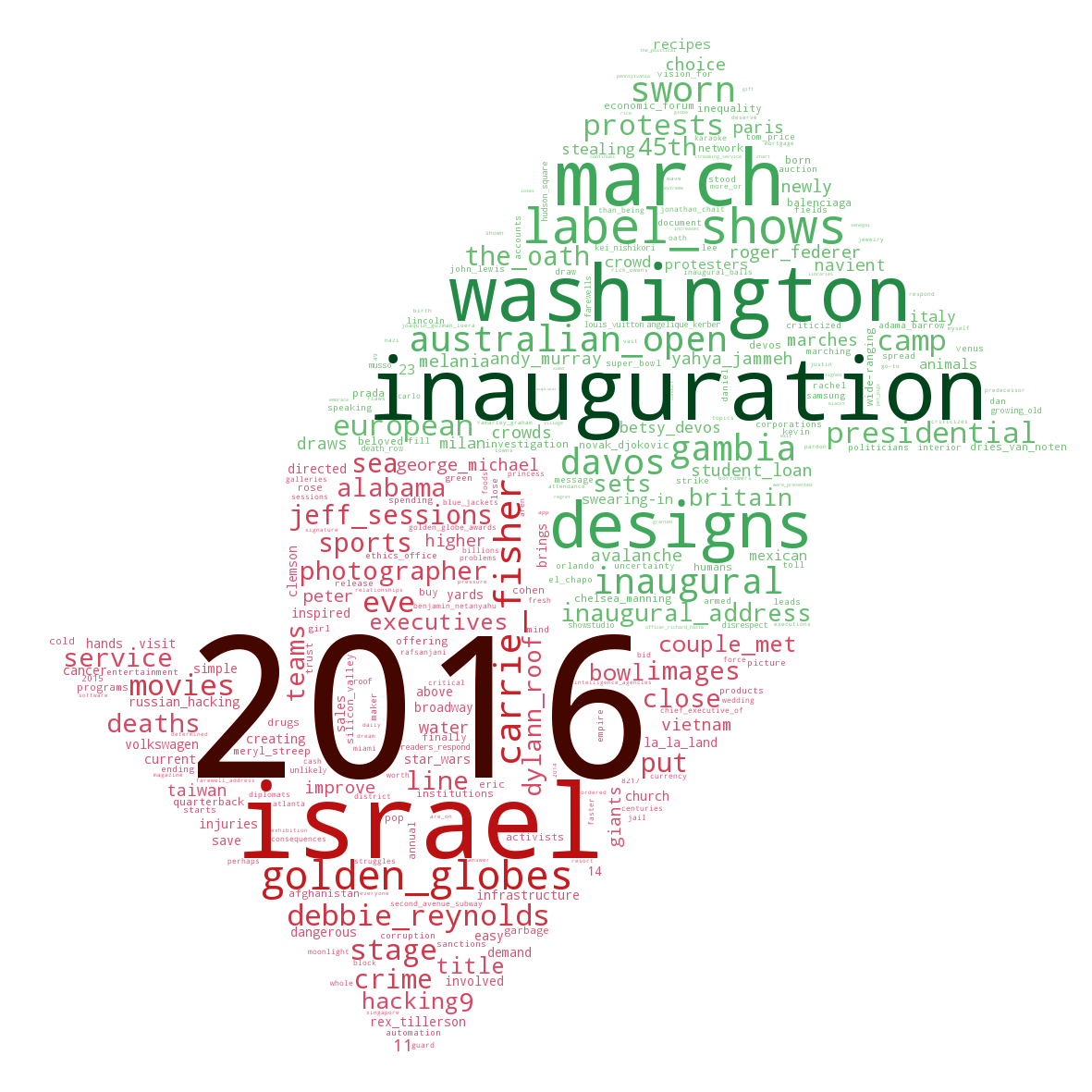}
  \caption{Relevant words in NY Times article snippets during the week of president Trump's inauguration (\emph{green/up}) and three weeks prior (\emph{red/down}).}
  \label{fig:distinctive}
\end{figure}
\begin{figure}[!h]
  \centering 
      \includegraphics[width=\columnwidth]{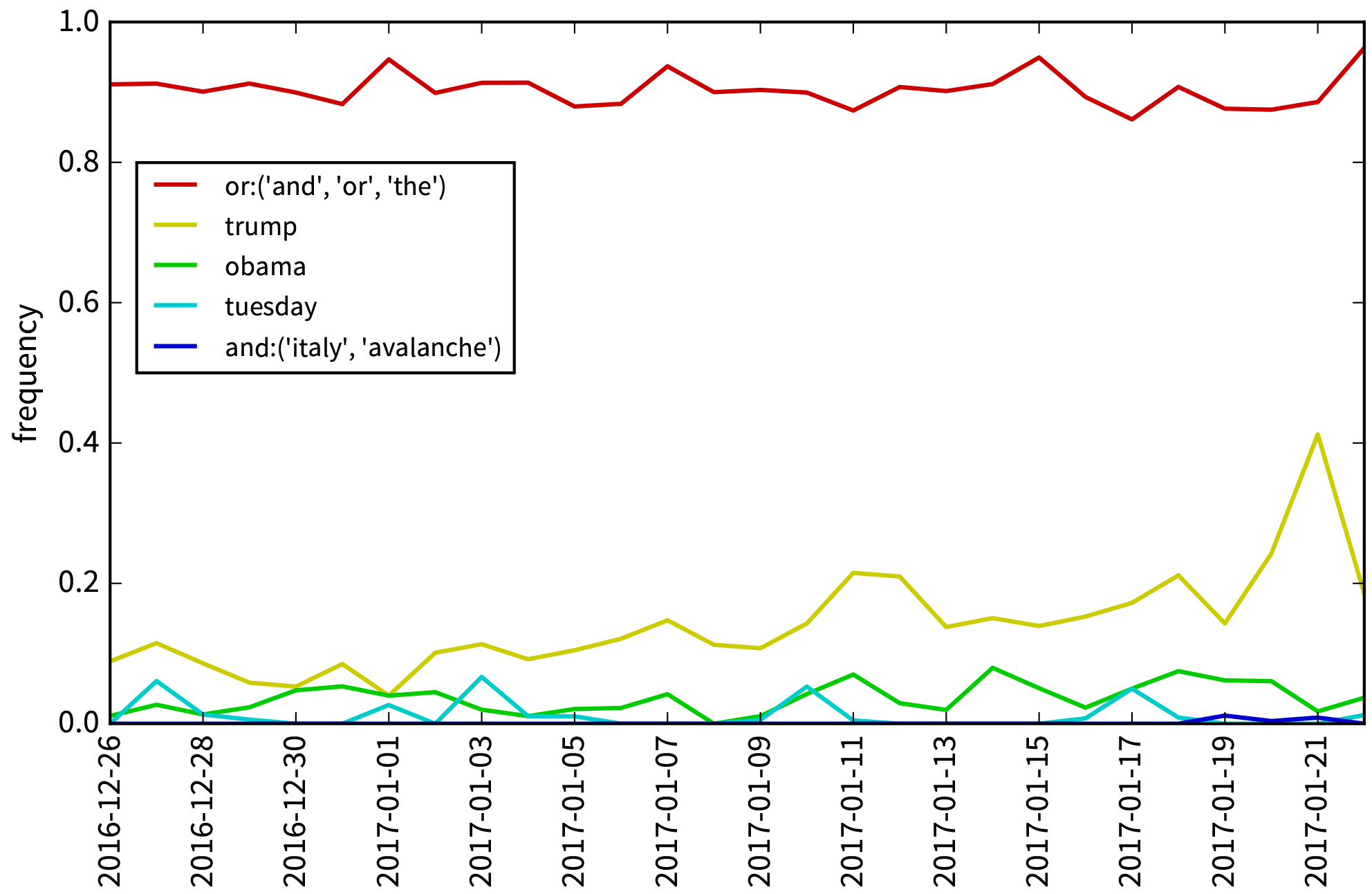}
  \caption{Frequencies of selected words in NY Times article snippets from different days.}
  \label{fig:nyt_occurrences}
\end{figure}
\begin{figure*}[!h]
  \centering 
      \includegraphics[width=0.3\textwidth]{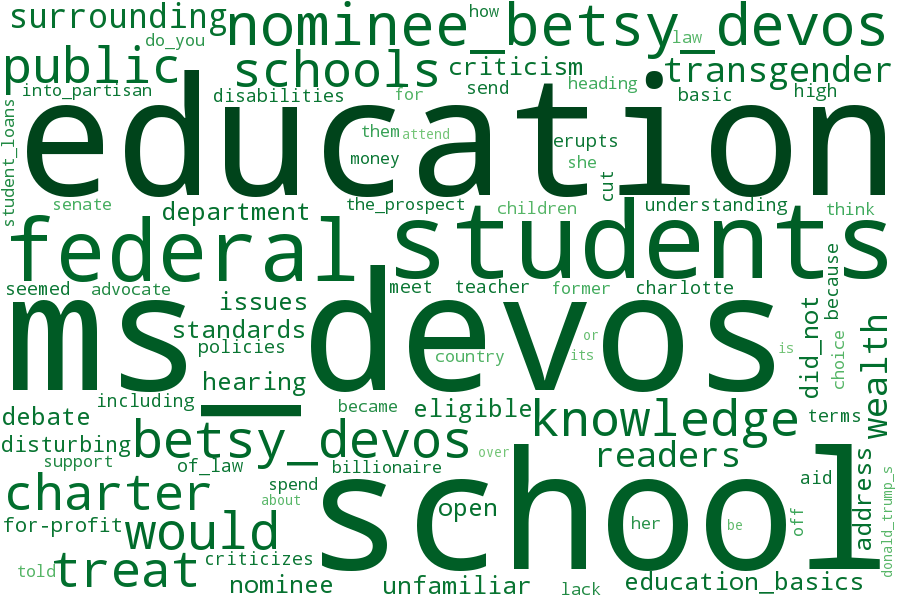}\hspace{0.32cm}
      \includegraphics[width=0.3\textwidth]{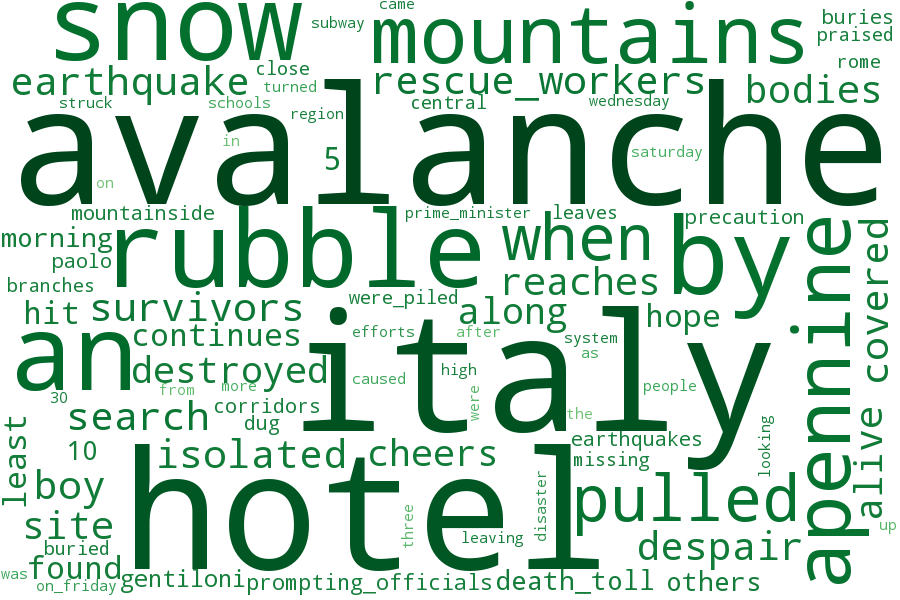}\\
      \includegraphics[width=0.65\textwidth]{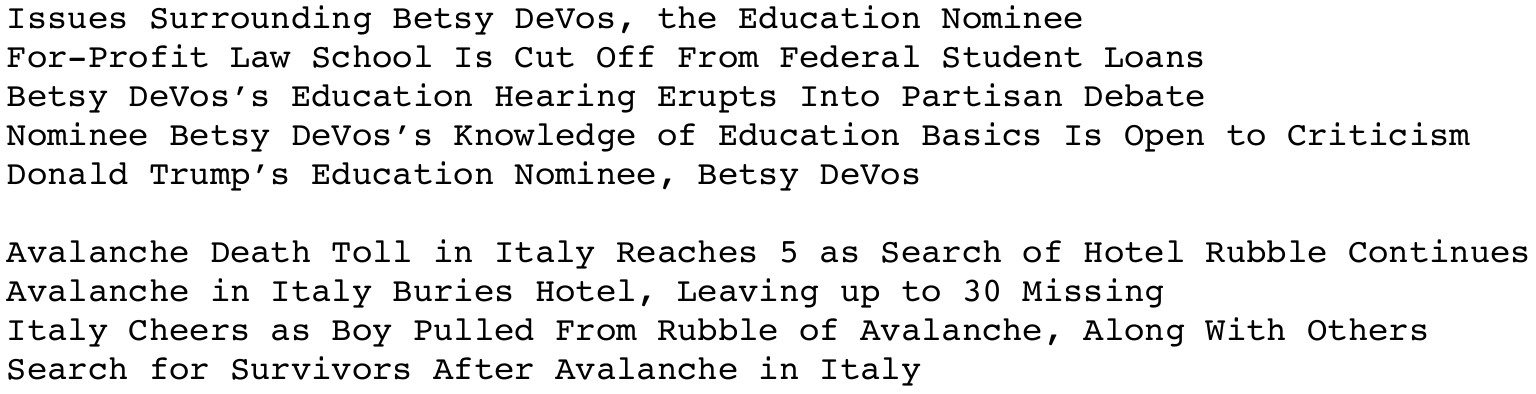}
  \caption{Word clouds created from the relevant words identified for two of over 50 clusters during the week Jan $16^{\text{th}}$-$22^{\text{nd}}$, 2017 and corresponding headlines.}
  \label{fig:wordcloud_clusters}
\end{figure*}

Before we cluster the texts, if we just manually split them into the articles published during the week of the inauguration ($c_1$) and before ($c_2$), the identified relevant words already reveals clear trends (Fig.~\ref{fig:distinctive}). Obviously, the main focus that week was on the inauguration itself, however it already becomes apparent that this will be followed by protest marches and also the Australian Open was happening at that time. When looking at the occurrence frequencies of different words over time (Fig.~\ref{fig:nyt_occurrences}), we can see the spike of `Trump' at the day of his inauguration, but while some stopwords occur equally frequent on all days, other rather meaningless words such as `Tuesday' have clear spikes as well (on Tuesdays). Therefore, care has to be taken when contrasting articles from different times when computing relevant words, as it could easily happen that these meaningless words are picked up as well simply because e.g.~one month contains more Tuesdays than another month used for comparison.

To identify trending topics, the articles from the week of the inauguration were clustered using DBSCAN. When enforcing a minimum cosine similarity of $0.55$ to other samples of a cluster as well as at least three articles per cluster, we obtain over 50 clusters for this week (as well as several articles considered `noise'). While some clusters correspond to specific sections of the newspaper (e.g.~corrections to articles published the days before), others indeed refer to meaningful events that happened that week, e.g.~the nomination of Betsy DeVos or an avalanche in Italy (Fig.~\ref{fig:wordcloud_clusters}). 

\section{Conclusion}
Examining the relevant words that summarize different groups of documents in a dataset is a very helpful step in the exploratory analysis of a collection of texts. It allows to quickly grasp the contents of documents belonging to certain clusters and can help identify salient topics, which is important if one is faced with a large dataset and quickly needs to find documents of interest.

We have explained how to compute a relevancy score for individual words depending on the number of documents in the target cluster this word occurs in compared to other clusters. This method is very fast and robust with respect to small or varying numbers of samples per cluster. The usefulness of our approach was demonstrated by using the obtained word clouds to identify trending topics in recent New York Times article snippets.

We hope the provided code will encourage other people faced with large collections of texts to quickly dive into the analysis and to thoroughly explore new datasets.

\section*{Acknowledgments}
We would like to thank Christoph Hartmann for his helpful comments on an earlier version of this manuscript. Franziska Horn acknowledges funding from the Elsa-Neumann scholarship from the TU Berlin.

\bibliography{../../phd_collected}

\begin{thebibliography}{16}
\providecommand{\natexlab}[1]{#1}
\providecommand{\url}[1]{\texttt{#1}}
\expandafter\ifx\csname urlstyle\endcsname\relax
  \providecommand{\doi}[1]{doi: #1}\else
  \providecommand{\doi}{doi: \begingroup \urlstyle{rm}\Url}\fi

\bibitem[Arras et~al.(2016{\natexlab{a}})Arras, Horn, Montavon, M{\"u}ller, and
  Samek]{arras2016explaining}
Leila Arras, Franziska Horn, Gr{\'e}goire Montavon, Klaus-Robert M{\"u}ller,
  and Wojciech Samek.
\newblock {Explaining Predictions of Non-Linear Classifiers in NLP}.
\newblock In \emph{Proceedings of the 1st Workshop on Representation Learning
  for NLP}, pages 1--7. Association for Computational Linguistics,
  2016{\natexlab{a}}.

\bibitem[Arras et~al.(2016{\natexlab{b}})Arras, Horn, Montavon, M{\"u}ller, and
  Samek]{arras2016relevant}
Leila Arras, Franziska Horn, Gr{\'e}goire Montavon, Klaus-Robert M{\"u}ller,
  and Wojciech Samek.
\newblock "what is relevant in a text document?": An interpretable machine
  learning approach.
\newblock \emph{arXiv preprint arXiv:1612.07843}, 2016{\natexlab{b}}.

\bibitem[Bach et~al.(2015)Bach, Binder, Montavon, Klauschen, M{\"u}ller, and
  Samek]{bach2015pixel}
Sebastian Bach, Alexander Binder, Gr{\'e}goire Montavon, Frederick Klauschen,
  Klaus-Robert M{\"u}ller, and Wojciech Samek.
\newblock On pixel-wise explanations for non-linear classifier decisions by
  layer-wise relevance propagation.
\newblock \emph{{PLOS ONE}}, 10\penalty0 (7):\penalty0 e0130140, 2015.

\bibitem[Ester et~al.(1996)Ester, Kriegel, Sander, Xu, et~al.]{ester1996dbscan}
Martin Ester, Hans-Peter Kriegel, J{\"o}rg Sander, Xiaowei Xu, et~al.
\newblock A density-based algorithm for discovering clusters in large spatial
  databases with noise.
\newblock In \emph{Kdd}, volume~96, pages 226--231, 1996.

\bibitem[Forman(2003)]{forman2003extensive}
George Forman.
\newblock An extensive empirical study of feature selection metrics for text
  classification.
\newblock \emph{The Journal of Machine Learning Research}, 3:\penalty0
  1289--1305, 2003.

\bibitem[Heimerl et~al.(2014)Heimerl, Lohmann, Lange, and
  Ertl]{heimerl2014word}
Florian Heimerl, Steffen Lohmann, Simon Lange, and Thomas Ertl.
\newblock Word cloud explorer: Text analytics based on word clouds.
\newblock In \emph{System Sciences (HICSS), 2014 47th Hawaii International
  Conference on}, pages 1833--1842. IEEE, 2014.

\bibitem[Horn et~al.(2017)Horn, Arras, Montavon, M{\"u}ller, and
  Samek]{horn2017exploring}
Franziska Horn, Leila Arras, Gr{\'e}goire Montavon, Klaus-Robert M{\"u}ller,
  and Wojciech Samek.
\newblock Exploring text datasets by visualizing relevant words.
\newblock \emph{arXiv preprint arXiv:1707.05261}, 2017.

\bibitem[Hulth(2003)]{hulth2003improved}
Anette Hulth.
\newblock Improved automatic keyword extraction given more linguistic
  knowledge.
\newblock In \emph{Proceedings of the 2003 conference on Empirical methods in
  natural language processing}, pages 216--223. Association for Computational
  Linguistics, 2003.

\bibitem[Lee and Kim(2008)]{lee2008news}
Sungjick Lee and Han-joon Kim.
\newblock News keyword extraction for topic tracking.
\newblock In \emph{Networked Computing and Advanced Information Management,
  2008. NCM'08. Fourth International Conference on}, volume~2, pages 554--559.
  IEEE, 2008.

\bibitem[Manning et~al.(2008)Manning, Raghavan, and Sch\"{u}tze]{irbook}
Christopher~D. Manning, Prabhakar Raghavan, and Hinrich Sch\"{u}tze.
\newblock \emph{Introduction to Information Retrieval}.
\newblock Cambridge University Press, New York, NY, USA, 2008.
\newblock ISBN 0521865719, 9780521865715.

\bibitem[McNaught and Lam(2010)]{mcnaught2010using}
Carmel McNaught and Paul Lam.
\newblock Using wordle as a supplementary research tool.
\newblock \emph{The qualitative report}, 15\penalty0 (3):\penalty0 630, 2010.

\bibitem[Mikolov et~al.(2013)Mikolov, Sutskever, Chen, Corrado, and
  Dean]{mikolov2013distributed}
Tomas Mikolov, Ilya Sutskever, Kai Chen, Greg~S Corrado, and Jeff Dean.
\newblock Distributed representations of words and phrases and their
  compositionality.
\newblock In \emph{Advances in neural information processing systems}, pages
  3111--3119, 2013.

\bibitem[Montavon et~al.(2017)Montavon, Samek, and
  M{\"u}ller]{montavon2017methods}
Gr{\'e}goire Montavon, Wojciech Samek, and Klaus-Robert M{\"u}ller.
\newblock Methods for interpreting and understanding deep neural networks.
\newblock \emph{arXiv preprint arXiv:1706.07979}, 2017.

\bibitem[Sch{\"o}lkopf et~al.(1998)Sch{\"o}lkopf, Smola, and
  M{\"u}ller]{scholkopf1998nonlinear}
Bernhard Sch{\"o}lkopf, Alexander Smola, and Klaus-Robert M{\"u}ller.
\newblock Nonlinear component analysis as a kernel eigenvalue problem.
\newblock \emph{Neural computation}, 10\penalty0 (5):\penalty0 1299--1319,
  1998.

\bibitem[Yang and Pedersen(1997)]{yang1997comparative}
Yiming Yang and Jan~O. Pedersen.
\newblock A comparative study on feature selection in text categorization.
\newblock In \emph{Proceedings of the Fourteenth International Conference on
  Machine Learning}, ICML '97, pages 412--420, San Francisco, CA, USA, 1997.
  Morgan Kaufmann Publishers Inc.
\newblock ISBN 1-55860-486-3.

\bibitem[Zhang et~al.(2006)Zhang, Xu, Tang, and Li]{zhang2006keyword}
Kuo Zhang, Hui Xu, Jie Tang, and Juanzi Li.
\newblock \emph{Keyword Extraction Using Support Vector Machine}, pages 85--96.
\newblock Springer Berlin Heidelberg, Berlin, Heidelberg, 2006.

\end{thebibliography}
\bibliographystyle{plainnat}

\end{document}